\documentclass{SCIS2019}

\begin{document}


\ArticleType{NEWS \& VIEWS}
\Year{2019}
\Month{}
\Vol{}
\No{}
\DOI{}
\ArtNo{}
\ReceiveDate{}
\ReviseDate{}
\AcceptDate{}
\OnlineDate{}

\title{CircuitNet: An Open-Source Dataset for Machine
Learning Applications in Electronic Design Automation (EDA)}{CircuitNet: An Open-Source Dataset for Machine Learning Applications in Electronic Design Automation (EDA)}

\author[1, 2]{Zhuomin Chai}{}
\author[1]{Yuxiang Zhao}{}
\author[1]{Yibo Lin}{yibolin@pku.edu.cn}
\author[2]{Wei Liu}{}
\author[1]{Runsheng Wang}{}
\author[1]{Ru Huang}{}

\AuthorMark{Zhuomin Chai}

\AuthorCitation{Zhuomin Chai, Yuxiang Zhao, Yibo Lin, et al}


\address[1]{School of Integrated Circuits, Peking Univeristy, Beijing {\rm 100871}, China}
\address[2]{School of Physics and Technology, Wuhan {\rm 430072}, China}

\maketitle

\begin{multicols}{2}
\noindent 
  The electronic design automation (EDA) community has been actively exploring machine learning (ML) for very large-scale integrated computer-aided design (VLSI CAD). Many studies explored learning-based techniques for cross-stage prediction tasks in the design flow to achieve faster design convergence. Although building ML models usually requires a large amount of data, most studies can only generate small internal datasets for validation because of the lack of large public datasets.
  In this essay, we present the first open-source dataset called \texttt{CircuitNet} for ML tasks in VLSI CAD. 

\begin{figure*}[tb]
\begin{minipage}[b]{.5\linewidth}
\centering
\resizebox*{0.97\linewidth}{0.57\linewidth}{
\begin{tabular}{|cccc||cc|}
\hline
\multicolumn{1}{|c|}{\multirow{2}{*}{Design}} & \multicolumn{3}{c||}{Netlist Statistics} & \multicolumn{2}{c|}{Synthesis Variations} \\ \cline{2-6}
\multicolumn{1}{|c|}{} & \multicolumn{1}{c|}{\#Cells} & \multicolumn{1}{c|}{\#Nets} & \multicolumn{1}{c||}{\begin{tabular}[c]{@{}c@{}}Cell Area\\ ($\mu m^2$)\end{tabular}} & \multicolumn{1}{c|}{\#Macros} & \begin{tabular}[c]{@{}c@{}}Frequency\\ (MHz)\end{tabular} \\ \hline
\multicolumn{1}{|c|}{RISCY-a} & \multicolumn{1}{c|}{44836} & \multicolumn{1}{c|}{80287} & \multicolumn{1}{c||}{65739} & \multicolumn{1}{c|}{\multirow{3}{*}{3/4/5}} & \multirow{6}{*}{50/200/500} \\ \cline{1-1}
\multicolumn{1}{|c|}{RISCY-FPU-a} & \multicolumn{1}{c|}{61677} & \multicolumn{1}{c|}{106429} & \multicolumn{1}{c||}{75985} & \multicolumn{1}{c|}{} &  \\ \cline{1-1}
\multicolumn{1}{|c|}{zero-riscy-a} & \multicolumn{1}{c|}{35017} & \multicolumn{1}{c|}{67472} & \multicolumn{1}{c||}{58631} & \multicolumn{1}{c|}{} &  \\ \cline{1-1} \cline{5-5}
\multicolumn{1}{|c|}{RISCY-b} & \multicolumn{1}{c|}{30207} & \multicolumn{1}{c|}{58452} & \multicolumn{1}{c||}{69779} & \multicolumn{1}{c|}{\multirow{3}{*}{13/14/15}} &  \\ \cline{1-1}
\multicolumn{1}{|c|}{RISCY-FPU-b} & \multicolumn{1}{c|}{47130} & \multicolumn{1}{c|}{84676} & \multicolumn{1}{c||}{80030} & \multicolumn{1}{c|}{} &  \\ \cline{1-1}
\multicolumn{1}{|c|}{zero-riscy-b} & \multicolumn{1}{c|}{20350} & \multicolumn{1}{c|}{45599} & \multicolumn{1}{c||}{62648} & \multicolumn{1}{c|}{} &  \\ \hline \hline
\multicolumn{6}{|c|}{Physical Design Variations} \\ \hline
\multicolumn{1}{|c|}{\begin{tabular}[c]{@{}c@{}}Utilizations\\ (\%)\end{tabular}} & \multicolumn{2}{c|}{\begin{tabular}[c]{@{}c@{}}\#Macro\\ Placement\end{tabular}} & \multicolumn{1}{c|}{\begin{tabular}[c]{@{}c@{}}\#Power Mesh\\ Setting\end{tabular}} & \multicolumn{2}{c|}{Filler Insertion} \\ \hline
\multicolumn{1}{|c|}{70/75/80/85/90} & \multicolumn{2}{c|}{3} & \multicolumn{1}{c|}{8} & \multicolumn{2}{c|}{\begin{tabular}[c]{@{}c@{}}After Placement\\ /After Routing\end{tabular}} \\ \hline
\end{tabular}
}
\caption*{(a)}
\end{minipage}%
\begin{minipage}[b]{.5\linewidth}
\centering
\includegraphics[width=1\textwidth]{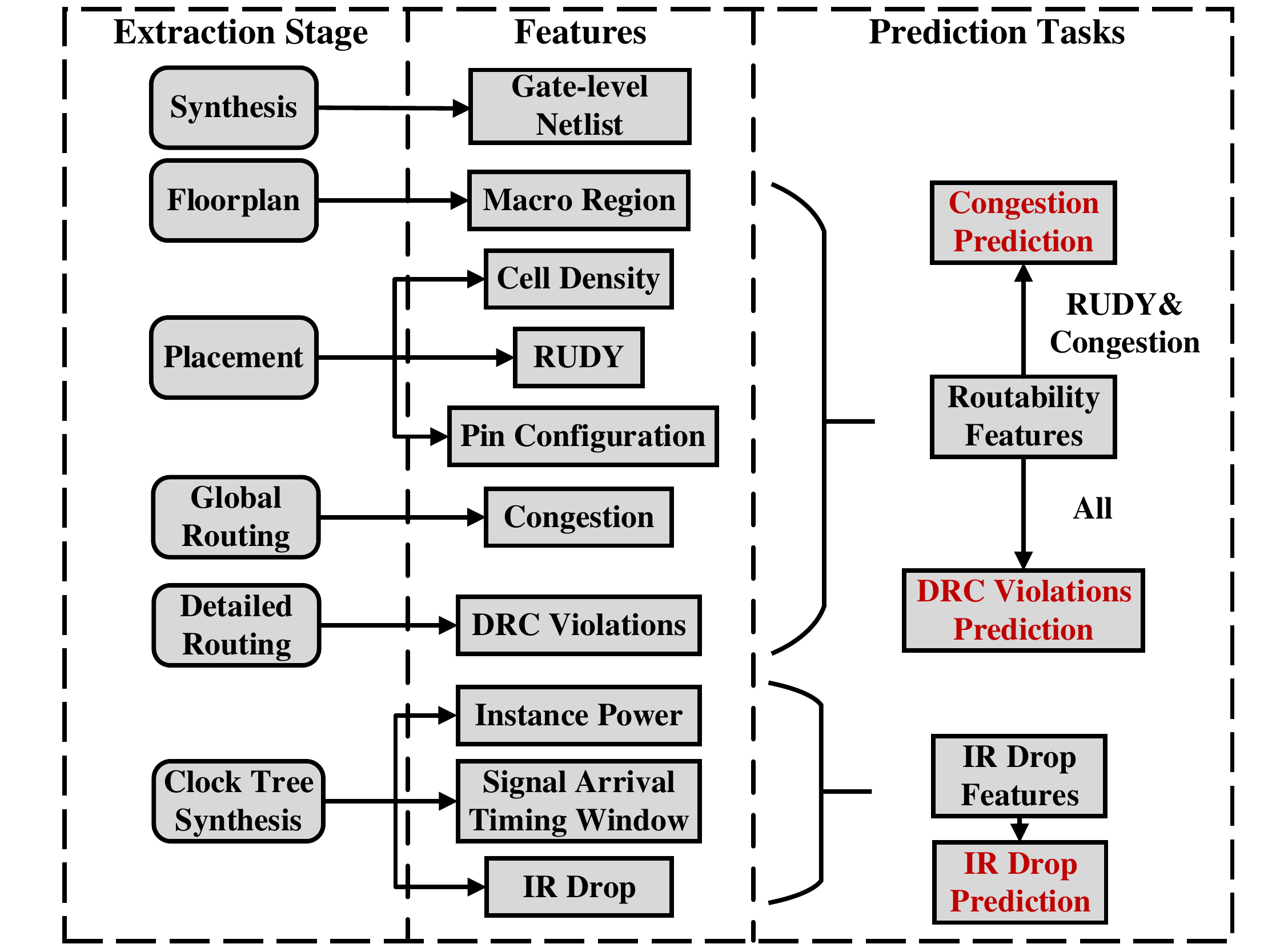}
\caption*{(b)}
\end{minipage}

\caption{(a) Statistics of designs and variations introduced during data collection. (b) Available features, their extraction stages, and prediction tasks in the experiments.}
\label{fig1}
\end{figure*}

\lettersection{Introduction} 
  VLSI circuit design can be divided into front-end design and back-end design. The front-end design implements the functionality of the circuit, and then, the back-end design transforms the circuit into manufacturable geometries, i.e., layouts.
  In advanced technology nodes, the back-end design is time-consuming because of iterative information feed-forward and feed-backward between the design stages during optimization.
  To accelerate this process, cross-stage prediction was introduced to replace the original long feedback loops between design stages with local loops within the design stages.
  As a promising method for fast and accurate cross-stage prediction, ML has been explored for various early-stage prediction tasks in the design flow, including routability and IR drop~\cite{1}. 

  Despite the active research on ML for CAD, there remain some challenges in this field. There is almost no public dataset dedicated to ML for CAD applications because of license restrictions and domain-specific expertise for data generation. Meanwhile, the existing datasets obtained from CAD contests are often incomplete and not designed for ML applications~\cite{2}. 
  The lack of public datasets raises challenges such as difficulty in benchmarking and reproducing previous work, limited research scope from limited data access, and a high bar for new researchers, which slows down further advancements in this field. To this end, we present the first open-source dataset, \texttt{CircuitNet}, which provides holistic support for cross-stage prediction tasks in back-end design with diverse samples.

\lettersection{Dataset overview} 
  The statistics of the dataset are summarized in Figure~\ref{fig1}. 
  We followed two steps to generate the dataset: data collection and feature extraction.  

  Data collection consisted of two stages: logic synthesis and physical design. 
  In logic synthesis, the RISC-V designs were mapped from register transfer level (RTL) designs to gate-level netlists in the 28 nm technology node with Synopsys Design Compiler. 
  Then, the physical design transformed the netlists into layouts with Cadence Innovus. 
  We improved the diversity of the dataset by introducing different settings in logic synthesis and physical design, as shown in Figure~\ref{fig1}(a). 
  These settings contributed to variations in utilization, routing resources, macro locations, etc., reflecting diverse situations in the back-end design flow.
  Each design has 2160 settings, and all the designs have 12,960 runs of the back-end design flow. 
  Eventually, we obtained 10,242 layouts after excluding the failed runs.

  In feature extraction, features were extracted at various design stages to support different cross-stage prediction tasks, as shown in Figure~\ref{fig1}(b). 
  We included both graph-like features (i.e., gate-level netlists) and image-like features (i.e., two-dimensional feature maps extracted from the physical layouts, as the design information can be naturally represented by image-like data by dividing a layout into tiles and regarding each tile as a pixel). 
  These features are widely adopted in the state-of-the-art (SOTA) routability and IR drop prediction models~\cite{3,4,5}. 

                                                                                                                                  
\lettersection{Dataset evaluation} 
  To evaluate the effectiveness of CircuitNet, we further conducted experiments on three prediction tasks: congestion, design rule check (DRC) violation, and IR drop. 
  Each experiment adopted a method from recent studies~\cite{3,4,5} and evaluated its result on CircuitNet with the same evaluation metrics as in the original studies. 
  These methods utilized image-like features to train a generative model, such as fully convolutional networks (FCNs) and U-Net, that formulating the prediction task into an image-to-image translation task.
  A detailed manual about the setup of these experiments is available on our webpage. Herein, we briefly introduce our results.

  First, for congestion prediction, we used the normalized root-mean-square-error and structural similarity index measure as metrics to evaluate pixel-level accuracy. The corresponding results for an FCN based method~\cite{3} were 0.040 and 0.80, respectively. 
  Second, for DRC violation prediction, we considered the area under the curve (AUC) of the receiver operating characteristic (ROC) curve and that of the precision-recall (PR) curve as the metrics for imbalanced learning. The corresponding results for an FCN based method~\cite{4} were 0.95 and 0.63, respectively. 
  Finally, for IR drop prediction, we evaluated the AUC of the ROC curve and that of the PR curve. The corresponding results for a U-Net based method~\cite{5} were 0.94 and 0.83.
  Overall, our results are relatively consistent with the original publications and demonstrate the effectiveness of CircuitNet. 

\lettersection{Usage} 
  We separated the features shown in Figure~\ref{fig1}(b) and stored them in different directories to enable custom applications. 
  We provided scripts for preprocessing and combining different features for training and testing used in the above experiments as references.

\lettersection{Access methods}
The user guide and the download link for CircuitNet can be accessed from \url{https://circuitnet.github.io}.





\end{multicols}

\end{document}